%% file: Sage_LaTeX_Guidelines.tex
\documentclass[Afour,sagev,times]{sagej}

\usepackage{moreverb,url}

\usepackage[colorlinks,bookmarksopen,bookmarksnumbered,citecolor=red,urlcolor=red]{hyperref}
\usepackage[T1]{fontenc}
\usepackage[utf8]{inputenc}
\usepackage{tabularx}

\usepackage{algorithm}
\usepackage{algorithmic}

\newcommand\BibTeX{{\rmfamily B\kern-.05em \textsc{i\kern-.025em b}\kern-.08em
T\kern-.1667em\lower.7ex\hbox{E}\kern-.125emX}}

\def\volumeyear{2016}

\begin{document}

\runninghead{Steinmetz et al.}

\title{World Models for Autonomous Navigation of Terrestrial Robots from LIDAR observations}

\author{Raul Steinmetz\affilnum{1, 2}, Fabio Demo Rosa\affilnum{1},  Victor Augusto Kich\affilnum{2},  Jair Augusto Bottega\affilnum{2}, Ricardo Bedin Grando\affilnum{3, 4} and Daniel Fernando Tello Gamarra\affilnum{1}}

\affiliation{
\affilnum{1} Universidade Federal de Santa Maria, Brazil\\
\affilnum{2} University of Tsukuba, Japan\\
\affilnum{3} Universidade Federal de Rio Grande\\
\affilnum{4} Universidad Tecnológica del Uruguay, Uruguay\\
}

\corrauth{Daniel Fernando Tello Gamarra}

\email{daniel.gamarra@ufsm.br}

\input{sessions/0_abstract}

\maketitle

\input{sessions/1_introduction}
\input{sessions/2_related}
\input{sessions/3_theoretical}
\input{sessions/4_methodology}
\input{sessions/5_results}
\input{sessions/6_conclusion}

\input{Sage_LaTeX_Guidelines.bbl}
\end{document}

%% file: sessions/0_abstract.tex
\begin{abstract}
Autonomous navigation of terrestrial robots using Reinforcement Learning (RL) from LIDAR observations remains challenging due to the high dimensionality of sensor data and the sample inefficiency of model-free approaches. Conventional policy networks struggle to process full-resolution LIDAR inputs, forcing prior works to rely on simplified observations that reduce spatial awareness and navigation robustness. This paper presents a novel model-based RL framework built on top of the DreamerV3 algorithm, integrating a Multi-Layer Perceptron Variational Autoencoder (MLP-VAE) within a world model to encode high-dimensional LIDAR readings into compact latent representations. These latent features, combined with a learned dynamics predictor, enable efficient imagination-based policy optimization. Experiments on simulated TurtleBot3 navigation tasks demonstrate that the proposed architecture achieves faster convergence and higher success rate compared to model-free baselines such as SAC, DDPG, and TD3. It is worth emphasizing that the DreamerV3-based agent attains a 100\% success rate across all evaluated environments when using the full dataset of the Turtlebot3 LIDAR (360 readings), while model-free methods plateaued below 85\%. These findings demonstrate that integrating predictive world models with learned latent representations enables more efficient and robust navigation from high-dimensional sensory data.

\end{abstract}

\keywords{Deep Reinforcement Learning, Autonomous Navigation, Terrestrial Mobile Robot, TurtleBot3, World Models}

%% file: sessions/1_introduction.tex
\section*{Supplementary Material}

The code and data used in this study are publicly available at: \href{https://github.com/raulsteinmetz/turtlebot-dreamerv3}{https://github.com/raulsteinmetz/turtlebot-dreamerv3}.

\section{Introduction}
Autonomous navigation of terrestrial robots has numerous practical applications, including space exploration~\cite{seeni2010robot}, mining operations~\cite{martinez2020mechatronics}, agriculture~\cite{grassi2018application}, household tasks~\cite{fiorini2000cleaning}, and industrial environments~\cite{keith2024review}. Deep Reinforcement Learning (DRL)~\cite{sutton2018reinforcement} has emerged as a powerful approach to enabling robots to autonomously learn complex behaviors, dynamically adapting to diverse environments through interactions and feedback~\cite{zhu2021deep}. DRL algorithms have demonstrated significant potential, offering adaptive solutions to robot navigation problems.

Distance sensors, especially Light Detection and Ranging (LIDAR), are widely employed in mapless DRL-based navigation tasks due to their reliability, computational simplicity, and consistency across simulation and real-world deployment. The TurtleBot3~\cite{turtlebot3} robot is widely used as a benchmark platform for evaluating DRL methods in mobile navigation from LIDAR sensor observations. Prior work has applied discrete-action algorithms, such as Deep Q-Network (DQN)~\cite{escobar2023autonomous}, Double Deep Q-Network (DDQN)~\cite{de2022double}, and State-Action-Reward-State-Action (SARSA)~\cite{anas2021comparison}, as well as continuous-action algorithms like Soft Actor Critic (SAC)~\cite{de2021soft}, Deep Deterministic Policy Gradient (DDPG)~\cite{grando2022deterministic}, and Twin Delayed DDPG (TD3)~\cite{li2022path}. 

Model-free continuous-control algorithms such as SAC~\cite{haarnoja2018soft}, DDPG~\cite{lillicrap2015continuous}, and TD3~\cite{fujimoto2018addressing} have achieved satisfactory performance in relatively simple environments. However, their dependence on direct interaction with the environment leads to prolonged training periods and inefficient sample utilization. Furthermore, these methods generally process LIDAR sensor readings directly within policy networks using linear layers, which works adequately for a limited number of sensor inputs (usually fewer than 20 readings~\cite{de2021soft, grando2022deterministic, li2022path}). When handling extensive sensor arrays, such as the complete set of 360 readings from the LIDAR sensor of the TurtleBot3, this direct processing approach struggles due to increased representation complexity and sparse reward signals. These challenges complicate the extraction of meaningful features, hinder gradient informativeness, and degrade policy training, leading to higher failure rates even in basic scenarios.

In contrast, model-based DRL approaches explicitly build an internal predictive model of the environment, anticipating future states and rewards~\cite{moerland2023model}. This predictive capability significantly enhances decision-making efficiency, requiring fewer interactions with the actual environment and substantially improving sample efficiency and reducing training time~\cite{janner2019trust}. Additionally, recent studies indicate that encoding sensor observations into compact, lower-dimensional representations using unsupervised learning can significantly enhance DRL performance compared to directly feeding raw data into policy networks~\cite{zheng2024texttt}. Such encodings allow efficient processing of richer sensor inputs and can lead to more robust and effective navigation policies.

Although model-free reinforcement learning methods remain widely used, recent years have shown a strong shift of the state-of-the-art in reinforcement learning towards model-based approaches. Algorithms such as DreamerV3~\cite{hafner2023mastering} and TD-MPC2~\cite{hansen2023td} have outperformed model-free baselines across a wide range of established benchmarks, achieving higher sample efficiency and robustness. This demonstrates a clear trend in the RL community, where predictive models and imagination-based rollouts provide many advantages over purely reactive policies. Despite this progress, the application of model-based reinforcement learning to mobile robot navigation from LIDAR observations has not yet been explored. This gap motivates the present study: to investigate whether the proven benefits of model-based reinforcement learning can be realized in LIDAR-based mapless terrestrial navigation tasks.

At the same time, it is worth noting that research outside the DRL community continues to advance model-free control strategies in various continuous-control domains. For instance, smoothed functional algorithms have been employed to optimize Proportional–Integral–Derivative (PID) controllers for wind turbines~\cite{ahmad2025improved}, while norm-limited Simultaneous Perturbation Stochastic Approximation (SPSA) has been applied for active oscillation control in drivetrains with backlash nonlinearity~\cite{yonezawa2024experimental}. Other recent contributions include cost-based extremum-seeking methods for model-free stabilization~\cite{dubbioso2024model} and fractional-order PID controller designs optimized through heuristic search techniques~\cite{nataraj2025design}.

Additionally, several optimization-based approaches have been proposed for robotic systems that share conceptual similarities with model-free control. For example, a Particle Swarm Optimization-based neural network PID controller has been developed for four-wheeled omnidirectional robots to enhance trajectory tracking~\cite{al2023pso}, and a Bee Algorithm has been used to tune nonlinear PID variants for two-link robotic arms, achieving improved stability under load disturbances~\cite{kadhim2024bee}. More recently, the Pelican Optimization Algorithm has been applied to mobile robot trajectory planning, enabling collision-free navigation and reduced path error compared to classical methods~\cite{khaleel2024improved}. These works demonstrate the continued relevance of heuristic, model-free optimization techniques in robotics. However, such methods rely on explicit cost functions and lack predictive environmental modeling, distinguishing them fundamentally from the model-based DRL paradigm investigated in this study.

\begin{figure}[H]
    \centering
    \includegraphics[width=\columnwidth]{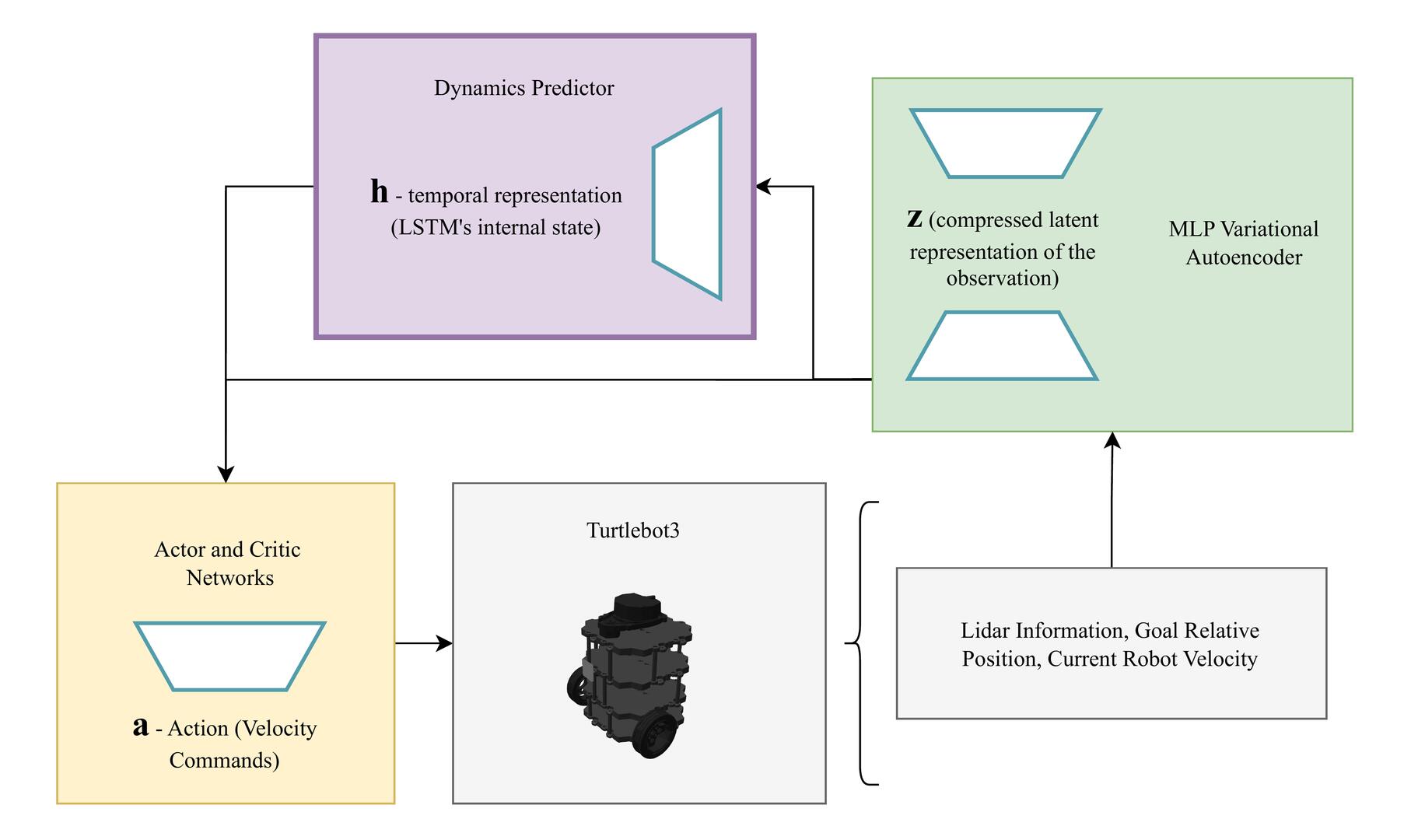}
    \caption{General diagram showing the different elements of the proposed agent architecture.}
    \label{fig:diagram}
\end{figure}

Motivated by these insights, this paper proposes a novel model-based DRL architecture for terrestrial robot navigation using LIDAR data. Built on top of the DreamerV3 algorithm~\cite{hafner2023mastering}, our architecture employs a Multi-Layer Perceptron Variational Autoencoder~\cite{kingma2013auto} to encode extensive LIDAR readings along with positional and velocity data into a compact latent representation. This encoded representation is integrated into the world model's dynamics predictor and controller, enabling efficient processing of large sensor arrays and enhancing decision-making capabilities. A diagram of the different elements of the architecture can be seen in Fig.~\ref{fig:diagram}.

The primary contributions of this study are:
\begin{enumerate}
\item A model-based deep reinforcement learning framework for terrestrial robot navigation:
Unlike prior model-free approaches that depend solely on reactive policy updates, the proposed world model explicitly learns environmental dynamics from LIDAR data. This predictive capability reduces the need for extensive environment interactions and improves learning efficiency and navigation success rates, addressing a long-standing limitation in mapless LIDAR-based navigation.
\item An encoder architecture enabling scalable processing of full 360-readings LIDAR observations:
By introducing a Multi-Layer Perceptron Variational Autoencoder to compress high-dimensional sensor data into compact latent representations, the method overcomes the input-size bottleneck that previously restricted DRL navigation to heavily reduced LIDAR subsets. This supports richer spatial perception.
\item Empirical validation of model-based DRL under complex navigation tasks:
Through systematic evaluation on the TurtleBot3 platform, the proposed architecture demonstrates superior convergence and stability compared to model-free baselines such as SAC, DDPG, and TD3. These results provide concrete experimental evidence of the advantages of predictive world modeling and variational autoencoders for continuous-control navigation.
\item An open implementation for research reproducibility:
The complete codebase and simulation setup are provided to enable independent validation and further methodological development within the scientific community.

\end{enumerate}

The paper is structured as follows: Section 2 reviews relevant literature. Section 3 describes the theoretical background underlying our proposed algorithm. Section 4 details the proposed methodology. Section 5 presents and discusses the experimental results. Finally, Section 6 concludes the paper with a discussion of the findings and suggestions for future research.

%% file: sessions/2_related.tex
\section{Related Works}\label{related_works}

This section reviews important prior work relevant to this paper. We focus specifically on research involving the TurtleBot3 navigation task, encoded representations in DRL, model-based DRL, World Models, and the Dreamer algorithm and its application in robotics.

\subsection{The TurtleBot3 Task}

The TurtleBot3 navigation task is a well-established benchmark for evaluating DRL algorithms on mobile navigation. In discrete control, Escobar-Naranjo et al.~\cite{escobar2023autonomous} proposed a framework based on the Deep Q-Network (DQN) algorithm. Similarly, De Moraes et al.~\cite{de2022double} compared DQN with the Double Deep Q-Network (DDQN) algorithm, finding that DDQN performs better in both simulated and real environments. Anas et al.~\cite{anas2021comparison} extended this comparison to include the State-Action-Reward-State-Action (SARSA) algorithm, further enriching insights into discrete-action methods for this task.

Continuous-control methods have also been explored extensively. De Jesus et al.~\cite{de2021soft} introduced a DRL method based on Soft Actor Critic (SAC), demonstrating superior performance for navigation tasks when using low-dimensional sensor data. Grando et al.~\cite{grando2022deterministic} conducted a detailed comparison between SAC and Deep Deterministic Policy Gradient (DDPG), analyzing various network architectures and their effects on learning efficiency. Additionally, Li et al.~\cite{li2022path} explored the Twin Delayed Deep Deterministic Policy Gradient (TD3) algorithm and its enhanced variant using prioritized memory replay.

However, these studies primarily rely on significantly reduced sets of LIDAR sensor readings to simplify the processing demands on the algorithms. While beneficial for computational simplicity, this reduction limits environmental perception, making it challenging to detect small or distant obstacles. Moreover, these algorithms employ model-free architectures, which do not leverage predictive dynamics models, potentially restricting their performance and extending training durations.

\subsection{Encoded Representations for Input Simplification}

Several studies have demonstrated that employing encoders to simplify input representations can substantially enhance the efficiency and performance of DRL algorithms. Zhang et al.~\cite{zhang2020learning} introduced latent representations to filter irrelevant information from observations, thereby focusing the policy's attention on task-relevant features.

Building upon this concept, Stooke et al.~\cite{stooke2021decoupling} demonstrated that separating representation learning from policy learning can outperform traditional end-to-end approaches, in which the same networks handle both observation processing and decision-making. By training encoders using unsupervised learning and independently optimizing policies on these latent representations, significant improvements in performance were observed across multiple benchmarks. Similarly, Prakash et al.~\cite{prakash2019use} showed that using autoencoders to compress high-dimensional data reduces computational demands and power consumption, particularly beneficial for embedded systems running DRL algorithms.

\subsection{Model-based DRL Approaches}

Model-based reinforcement learning has gained attention due to its sample efficiency and improved decision-making capabilities. Berkenkamp et al.~\cite{berkenkamp2017safe} addressed safety concerns during exploration, showing how models could help avoid unsafe states, thereby collecting safer and more informative data. Kaiser et al.~\cite{kaiser2019model} presented the Simulated Policy Learning algorithm, demonstrating that model-based methods can efficiently solve multiple Atari tasks with significantly fewer interactions compared to model-free counterparts.

In the robotics domain, model-based DRL methods have been successfully applied in various contexts. Thuruthel et al.~\cite{thuruthel2018model} used a predictive control algorithm for soft robotic manipulators, employing recurrent neural networks for dynamic modeling and trajectory optimization to guide policy decisions. Li et al.~\cite{li2020model} enhanced the DDPG algorithm by integrating an ensemble of deep neural networks to model system dynamics, significantly improving sample efficiency and control performance for simulated robotic manipulation tasks.

\subsection{World Models}

Building upon model-based reinforcement learning, Ha and Schmidhuber~\cite{ha2018world} introduced the concept of World Models. In this approach, recurrent neural networks create compressed spatial and temporal representations of environments, enabling agents to achieve exceptional performance and efficiency. Hafner et al.~\cite{hafner2019dream} further developed this idea with the Dreamer algorithm, which leverages latent imagination to solve long-horizon tasks from image-based inputs.

Dreamer was subsequently enhanced by Hafner et al. with the DreamerV2 algorithm~\cite{hafner2020mastering}, which operates within a compact latent space to achieve human-level performance on a wide array of Atari games. The framework was generalized further with DreamerV3~\cite{hafner2023mastering}, demonstrating versatility across numerous tasks with minimal hyperparameter tuning. Recently, Wu et al.~\cite{wu2023daydreamer} successfully applied Dreamer methods to physical robots operating in real-world environments, achieving mastery of complex locomotion tasks with limited real-world interactions.

To our knowledge, this paper is the first to apply a model-based DRL approach, specifically using the DreamerV3 framework, to the TurtleBot3 navigation task. Additionally, this work is the first to introduce encoded representations for processing extensive environmental observations, enabling the effective utilization of full LIDAR datasets for navigation tasks.

%% file: sessions/3_theoretical.tex
\section{Theoretical Background}\label{theoretical}

This section describes the foundational theoretical concepts and methodologies underlying this research. We first review model-free algorithms previously applied to this task, followed by explanations of model-based algorithms, world models, and the DreamerV3 framework.

\subsection{Reinforcement Learning Fundamentals}

Reinforcement Learning~\cite{sutton2018reinforcement} is a machine learning paradigm in which an agent learns to make decisions by interacting with an environment to maximize cumulative rewards. The agent operates in discrete time steps. At each step \( t \), it observes a state \( s_t \), executes an action \( a_t \), receives a reward \( r_t \), and transitions to a subsequent state \( s_{t+1} \). The objective is to learn a policy \( \pi(a|s) \) that maximizes the expected cumulative discounted reward:
\[
\mathbb{E}\left[\sum_{t=0}^{T} \gamma^t r_t\right]\text{~\cite{sutton2018reinforcement}},
\]
where \( \gamma \in [0, 1] \) is the discount factor balancing immediate and long-term rewards. 

\subsection{Model-Free Reinforcement Learning}

Model-free reinforcement learning algorithms directly learn optimal policies through interaction with the environment, without explicitly modeling environmental dynamics. In the context of continuous-action Turtlebot3 navigation, the algorithms previously used are Soft Actor-Critic (SAC)~\cite{haarnoja2018soft}, Deep Deterministic Policy Gradient (DDPG)~\cite{lillicrap2015continuous}, and Twin Delayed DDPG (TD3)~\cite{fujimoto2018addressing}.

SAC is an off-policy algorithm designed to maximize a trade-off between expected rewards and policy entropy, promoting exploration and preventing premature convergence to suboptimal policies. Its objective function is given by:
\[
\mathcal{J}(\pi) = \mathbb{E}_{\pi}\left[ \sum_{t=0}^{T}\gamma^t\left(r_t + \alpha \mathcal{H}(\pi(\cdot|s_t))\right)\right]\text{~\cite{haarnoja2018soft}},
\]
where \( \alpha \) controls the exploration-exploitation trade-off, and \( \mathcal{H}(\pi(\cdot|s_t)) \) denotes the policy entropy. SAC updates its policy using a reparameterization trick and employs a soft Q-function incorporating entropy to enhance exploration.

DDPG extends deterministic policy gradients to continuous action spaces using an actor-critic architecture. The actor network \( \mu(s|\theta^\mu) \) generates actions from states, and the critic network \( Q(s, a|\theta^Q) \) evaluates action values. The policy gradient is computed as:
\[
\nabla_{\theta^\mu} J \approx \mathbb{E}\left[\nabla_a Q(s, a|\theta^Q)|_{a=\mu(s)} \nabla_{\theta^\mu} \mu(s|\theta^\mu)\right]\text{~\cite{lillicrap2015continuous}}.
\]
While DDPG effectively handles continuous actions, it suffers from overestimation bias and training instability.

TD3 addresses DDPG’s limitations by employing clipped double-Q learning, delayed policy updates, and target policy smoothing. It utilizes two Q-networks, choosing the minimum estimate to mitigate overestimation. Policy updates occur less frequently to allow accurate value estimation, and noise is added to target policies to reduce variance. The TD3 target value is calculated as:
\[
y = r + \gamma \min_{i=1,2}Q_{\theta_i'}(s', \pi_{\phi'}(s'))\text{~\cite{fujimoto2018addressing}},
\]
where \( Q_{\theta_i'} \) and \( \pi_{\phi'} \) represent target Q-networks and policy, respectively, improving stability and performance.

\subsection{Model-Based Reinforcement Learning}

Model-based Reinforcement Learning (MBRL) explicitly learns a predictive model of environmental dynamics, enhancing learning efficiency by simulating interactions internally. This approach significantly reduces real-world interactions, as simulations using the learned model are typically faster.

In MBRL, neural networks predict subsequent states and rewards given current states and actions. Training the model typically involves minimizing mean squared error between predicted and actual outcomes. Once trained, simulated trajectories generated by the model facilitate efficient policy training and action selection.

\subsection{World Models}

World Models extend model-based RL by constructing generative neural network models that learn compressed spatial and temporal representations of the environment. These models consist of three primary components: the Vision (V) model, Memory (M) model, and Controller (C) model.

The Vision model employs a Variational Autoencoder (VAE)~\cite{kingma2013auto} to compress high-dimensional observations \( x \) into lower-dimensional latent vectors \( z \):
\[
z \sim q(z|x) = \mathcal{N}(\mu, \sigma^2I)\text{~\cite{kingma2013auto}}.
\]
The VAE optimizes the encoding to ensure accurate reconstruction of observations from the latent space.

The Memory model integrates a Long Short-Term Memory (LSTM) network and a Mixture Density Network (MDN), predicting future latent vectors \( z_{t+1} \) from the current action \( a_t \), latent vector \( z_t \), and hidden state \( h_t \):
\[
P(z_{t+1} | a_t, z_t, h_t) = \sum_{i=1}^{M}\pi_i\mathcal{N}(\mu_i, \sigma_i^2)\text{~\cite{ha2018world}}.
\]
The LSTM captures temporal dynamics, while the MDN manages predictive uncertainty.

The Controller model determines actions from latent vectors and hidden states through a linear mapping:
\[
a_t = W_c[z_t, h_t] + b_c\text{~\cite{ha2018world}},
\]
where \( W_c \) and \( b_c \) are model parameters. Training involves first optimizing the Vision and Memory models through random interactions, then training the Controller to maximize expected cumulative reward:
\[
R = \mathbb{E}\left[\sum_{t=0}^{T}\gamma^tr_t\right]\text{~\cite{sutton2018reinforcement}}.
\]
This policy can be trained fully on the World Model's understanding of the environment, and can be transferred to act on the actual environment after learning.

\subsection{DreamerV3}

DreamerV3 is a state-of-the-art RL algorithm that effectively generalizes across diverse tasks. It builds on world models, using a Recurrent State-Space Model (RSSM) that combines an LSTM and VAE to encode observations and predict future states robustly.

Key innovations of DreamerV3 include robust normalization via the symlog transformation:
\[
\text{symlog}(x) = \text{sign}(x)\cdot \log(1 + |x|)\text{~\cite{hafner2023mastering}},
\]
which provides numerical stability across varying value scales. It also applies return normalization, facilitating consistent reward scales for stable learning.

DreamerV3 trains policies through imagined rollouts generated by the world model, significantly enhancing sample efficiency. Additional improvements include discrete regression in the critic network, enhancing learning in sparse reward environments, and entropy regularization to promote exploration and avoid premature convergence. DreamerV3's scalability further reduces hyperparameter sensitivity, allowing consistent high performance across diverse tasks.

%% file: sessions/4_methodology.tex
\section{Methodology}\label{methodology}

This section describes the methodology employed in our research, including the framework for the TurtleBot3 environment and details of our proposed DRL architecture.

\subsection{The TurtleBot3 Framework}

The TurtleBot3 navigation task is formulated as a point-to-point challenge, where the robot navigates from a starting position to a randomly placed goal, avoiding collisions with obstacles. Each episode ends successfully if the robot reaches the target location, or unsuccessfully if it collides with an obstacle or exceeds the maximum step count.

Our simulation framework builds upon the approach of de Jesus et al.~\cite{de2021soft}, utilizing the Burger model of TurtleBot3. Their original ROS Noetic implementation was updated to ROS2, which offers enhanced performance, active community support, and long-term maintenance~\cite{ros2_changes}.

We also introduced an essential improvement in the reinforcement learning loop by controlling the rate of steps per second, which was previously unregulated. Lack of control in step frequency leads to several issues:

\begin{enumerate}
    \item \textbf{Hardware Dependence:} Algorithms trained on one type of hardware may perform poorly on different hardware due to variations in processing speed, notably impacting robots using low-power hardware.
    \item \textbf{Evaluation Consistency:} Inconsistent step rates between training (slower due to learning computations) and evaluation (faster due to no learning overhead) can negatively affect performance assessments.
    \item \textbf{Algorithm Comparability:} Different algorithms inherently have varied computation times, leading to unequal step rates and unfair performance comparisons.
\end{enumerate}

To mitigate these issues and simulate realistic operational conditions, we standardized the step rate at 6.67 steps per second (0.15 seconds per step). This frequency balances algorithmic computational needs, actuator responsiveness, and practical step limits for efficient navigation.

\begin{figure}[tbp]
    \centering
    \includegraphics[width=\columnwidth]{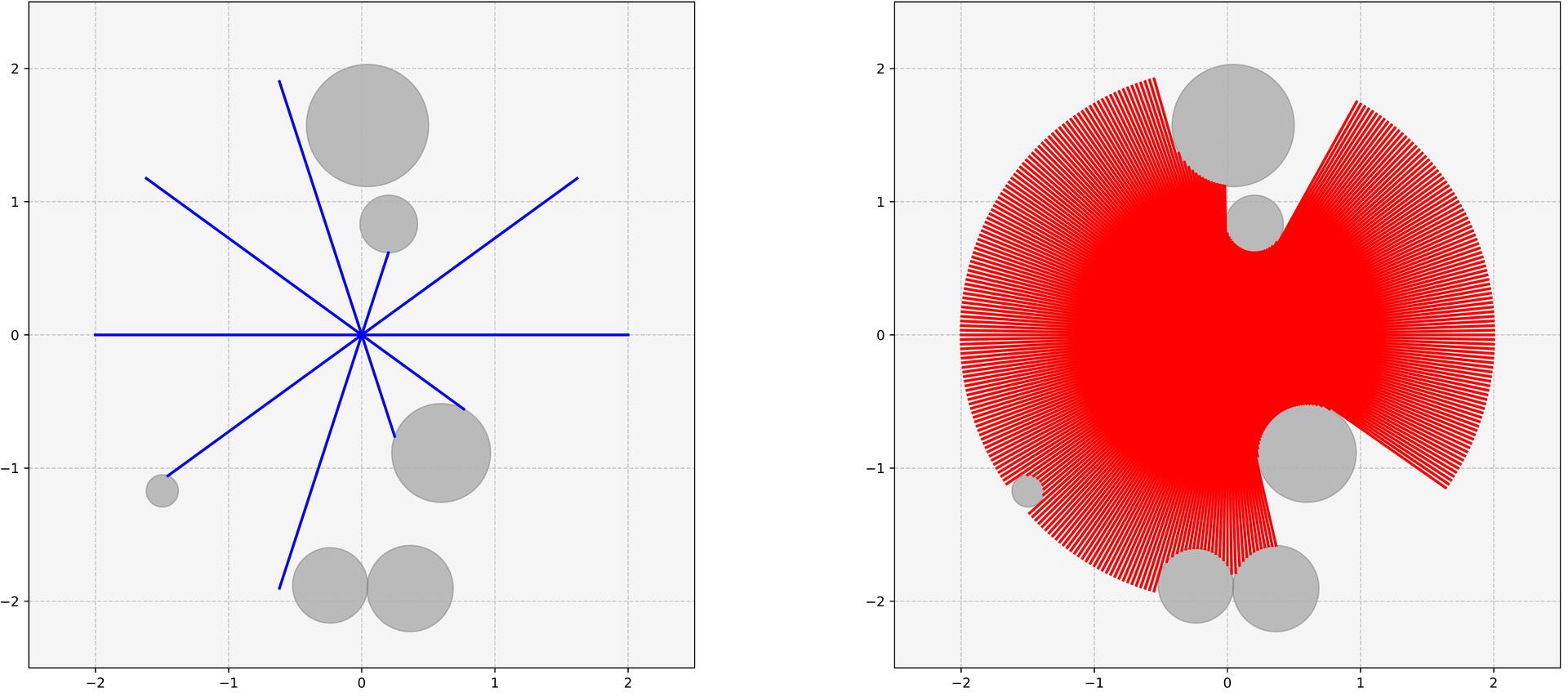}
    \caption{10 (blue) and 360 (red) distance readings description comparison.}
    \label{lidar}
\end{figure}

We tested two sensor configurations: the standard configuration with 10 equally spaced LIDAR readings every 36 degrees~\cite{de2021soft} , and the full sensor array of 360 readings (see Fig.~\ref{lidar}). While the reduced set offers computational efficiency and the possibility to process the data within small policy networks that learn on the reward signal, the complete set significantly improves environmental awareness, capturing smaller and distant objects that may otherwise be missed. In addition to LIDAR data, the input state includes the robot's relative position to the target (distance and angle) and its previous action (linear and angular velocities).

\begin{figure}[tbp]
    \centering
    \includegraphics[width=\columnwidth]{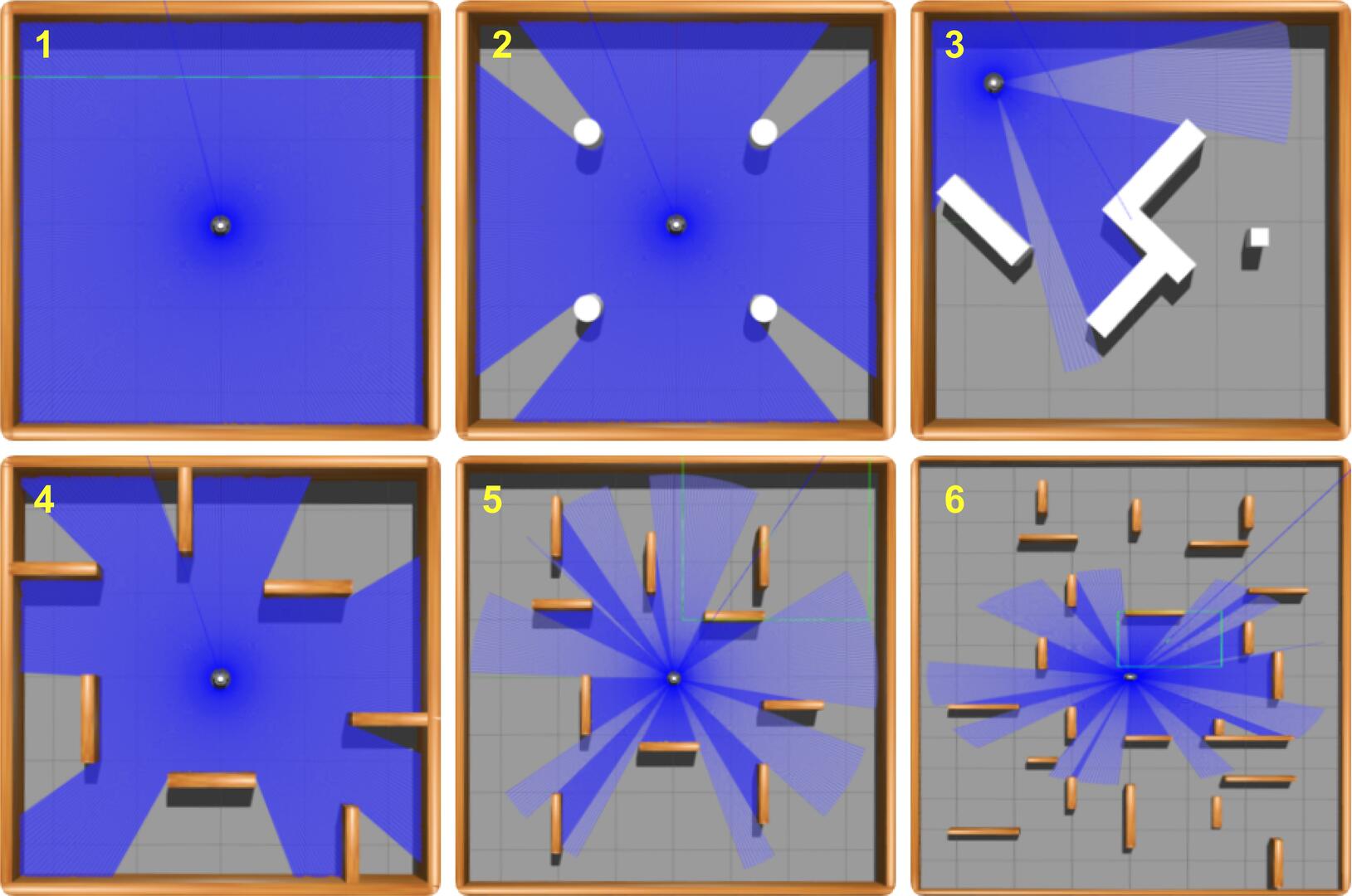}
    \caption{Environment setups used for algorithm comparison.}
    \label{stages}
\end{figure}

The environments used for training and evaluation are illustrated in Fig.~\ref{stages}. Stages 1–3 replicate setups from previous studies, while stages 4–6 introduce increasing complexity with more obstacles and larger navigation paths. Stage 4 is a dense 5x5 meter environment, stage 5 expands to 7.5x7.5 meters to test medium-distance navigation, and stage 6 further increases the challenge with a 7.5x14 meter space, testing long-distance navigation under sparse rewards.

The reward function \( r(s) \), dependent on time step \( t \), step limit \( T \), LIDAR readings \( D \), robot position \((x, y)\), and target position \((x', y')\), is defined as:
\[
r(s) = 
\begin{cases} 
100, & \text{if } \sqrt{(x - x')^2 + (y - y')^2} < 0.4 \text{ m}, \\
-10, & \text{if } \min(D) < 0.2 \text{ m or } t = T.
\end{cases}
\]

While alternative reward structures were considered, this sparse reward consistently yielded stable learning outcomes, aligning with prior findings in previous work~\cite{de2021soft, de2022double}.

Algorithm~\ref{alg:turtlebot_navigation} describes the reinforcement learning formulation for the TurtleBot3 navigation task. The robot observes its environment through LIDAR sensors, relative target position, and its own velocity. At each timestep, a policy network outputs continuous actions controlling the robot's linear and angular velocities. Episodes terminate successfully when the robot reaches the goal (reward $+100$), or unsuccessfully upon collision or timeout (reward $-10$). The control loop operates at $6.67$ Hz ($0.15$ seconds per step) to ensure consistent training and deployment across different hardware platforms.

\begin{algorithm*}[t]
\caption{TurtleBot3 Navigation: Reinforcement Learning Formulation}
\label{alg:turtlebot_navigation}
\begin{algorithmic}[1]
\STATE \textbf{Observation Space:} $o_t = [d_1, \ldots, d_N, \delta_t, \alpha_t, v_{t-1}^{\text{lin}}, v_{t-1}^{\text{ang}}]$ where $d_i \in [0, d_{\max}]$ are LIDAR readings, $\delta_t \in \mathbb{R}^+$ is distance to goal, $\alpha_t \in [-\pi, \pi]$ is angle to goal, and $N \in \{10, 360\}$
\STATE \textbf{Action Space:} $a_t = [v^{\text{lin}}, v^{\text{ang}}] \in [0, 1] \times [-1, 1]$ (normalized)
\STATE \textbf{Reward Function:} 
$r(o_t) = \begin{cases} 
r_{\text{success}} & \text{if } \delta_t < d_{\text{goal}} \text{ (success)} \\ 
r_{\text{fail}} & \text{if } \min(d_1, \ldots, d_N) < d_{\text{collision}} \text{ or } t \geq T_{\max} \text{ (collision/timeout)} \\ 
0 & \text{otherwise}
\end{cases}$
\STATE \textbf{Parameters:} $d_{\max} = 3.5$m, $d_{\text{goal}} = 0.4$m, $d_{\text{collision}} = 0.2$m, $r_{\text{success}} = +100$, $r_{\text{fail}} = -10$, $T_{\max} = 300$ or $500$, $\Delta t = 0.15$s
\STATE \textbf{Initialize:} Environment $\mathcal{E}$, policy network $\pi_\theta$, replay buffer $\mathcal{D}$
\FOR{$\text{episode} = 1$ to $N_{\text{episodes}}$}
    \STATE $(x_{\text{goal}}, y_{\text{goal}}) \gets$ sample random goal position
    \STATE $(x_0, y_0, \theta_0) \gets$ reset robot to initial pose
    \STATE $o_0 \gets$ \textsc{GetObservation}() \hfill $\triangleright$ Initial LIDAR, position, velocity
    \STATE $t \gets 0$, $\text{done} \gets \text{False}$
    \WHILE{not done}
        \STATE $a_t \gets \pi_\theta(o_t)$ \hfill $\triangleright$ Policy selects action
        \STATE \textit{// Environment interaction}
        \STATE Send velocity commands $(v^{\text{lin}}, v^{\text{ang}})$ to robot actuators
        \STATE \textit{// Perception}
        \STATE Read LIDAR: $\{d_1, \ldots, d_N\}$ from laser scanner
        \STATE Read odometry: $(x_{t+1}, y_{t+1}, \theta_{t+1})$ from wheel encoders
        \STATE \textit{// Observation construction}
        \STATE Compute distance: $\delta_{t+1} \gets \sqrt{(x_{\text{goal}} - x_{t+1})^2 + (y_{\text{goal}} - y_{t+1})^2}$
        \STATE Compute angle: $\alpha_{t+1} \gets \text{atan2}(y_{\text{goal}} - y_{t+1}, x_{\text{goal}} - x_{t+1}) - \theta_{t+1}$
        \STATE Build next observation: $o_{t+1} \gets [d_1, \ldots, d_N, \delta_{t+1}, \alpha_{t+1}, v^{\text{lin}}, v^{\text{ang}}]$
        \STATE \textit{// Reward computation and termination}
        \IF{$\delta_{t+1} < d_{\text{goal}}$}
            \STATE $r_t \gets r_{\text{success}}$, $\text{done} \gets \text{True}$ \hfill $\triangleright$ Goal reached
        \ELSIF{$\min(d_1, \ldots, d_N) < d_{\text{collision}}$ \textbf{or} $t \geq T_{\max}$}
            \STATE $r_t \gets r_{\text{fail}}$, $\text{done} \gets \text{True}$ \hfill $\triangleright$ Collision or timeout
        \ELSE
            \STATE $r_t \gets 0$, $\text{done} \gets \text{False}$ \hfill $\triangleright$ Continue navigation
        \ENDIF
        \STATE Store transition $(o_t, a_t, r_t, o_{t+1}, \text{done})$ in $\mathcal{D}$
        \STATE Update $\pi_\theta$
        \STATE $o_t \gets o_{t+1}$
        \STATE $t \gets t + 1$

        \STATE Wait for $\Delta t$ \hfill $\triangleright$ Maintain control rate
    \ENDWHILE
\ENDFOR
\end{algorithmic}
\end{algorithm*}

Our proposed architecture leverages the DreamerV3 framework, which incorporates a variational autoencoder (VAE) and a dynamics predictor for efficient handling of extensive sensor inputs and enhanced decision-making.

\begin{figure}[tbp]
    \centering
    \includegraphics[width=\columnwidth]{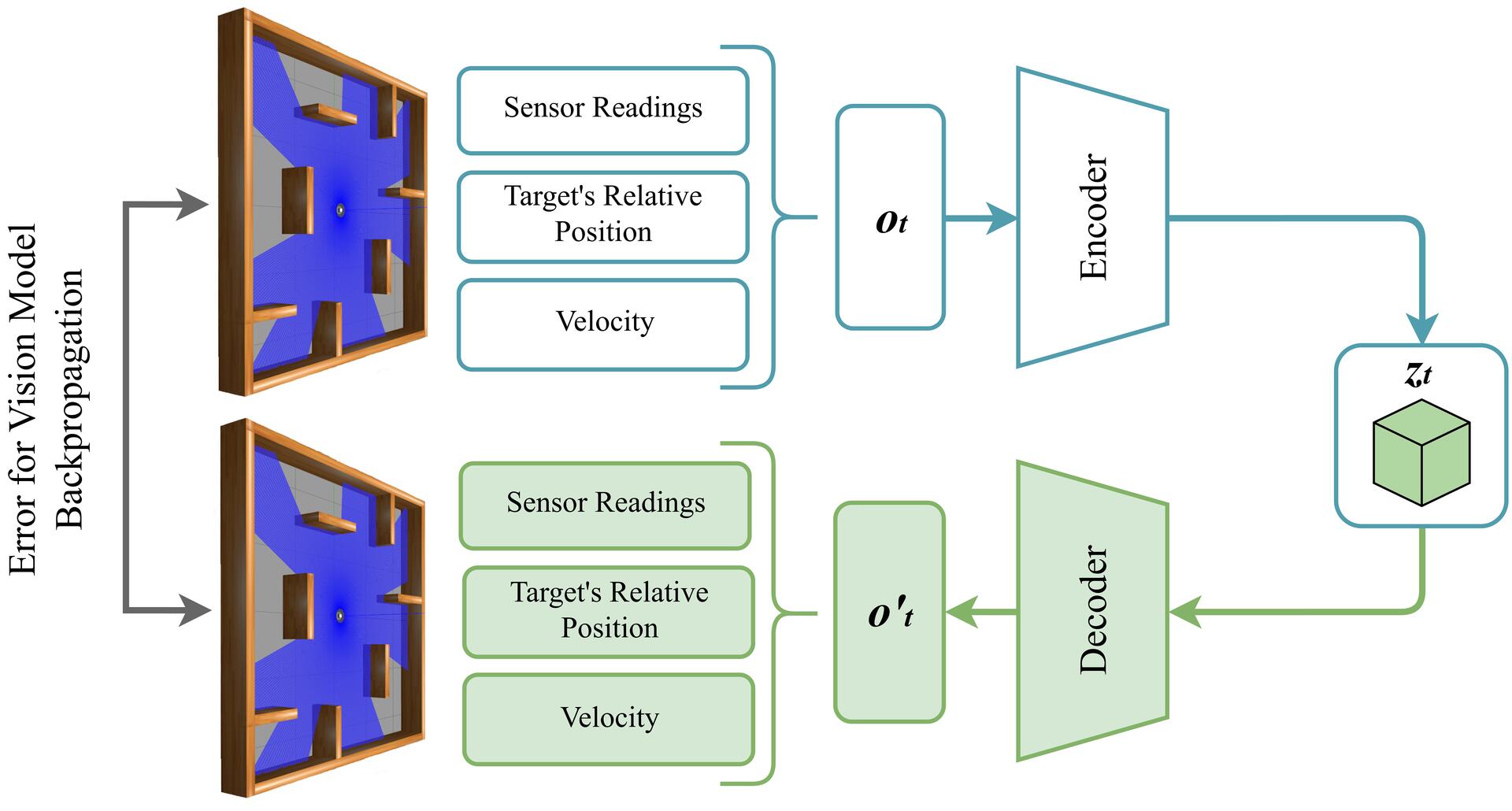}
    \caption{Vision model learning process diagram.}
    \label{fig:vision_model_learning}
\end{figure}

The Vision model is a Multilayer perceptron Variational Autoencoder (MLP-VAE), it compresses observations \( o_t \), comprising LIDAR readings, relative target position, and previous actions, into a latent representation \( z_t \). Training involves random environmental interactions, with the encoder mapping inputs to latent vectors and the decoder reconstructing the original observations. Reconstruction error (mean squared error) guides training, as depicted in Fig.~\ref{fig:vision_model_learning}.

\begin{figure}[tbp]
    \centering
    \includegraphics[width=\columnwidth]{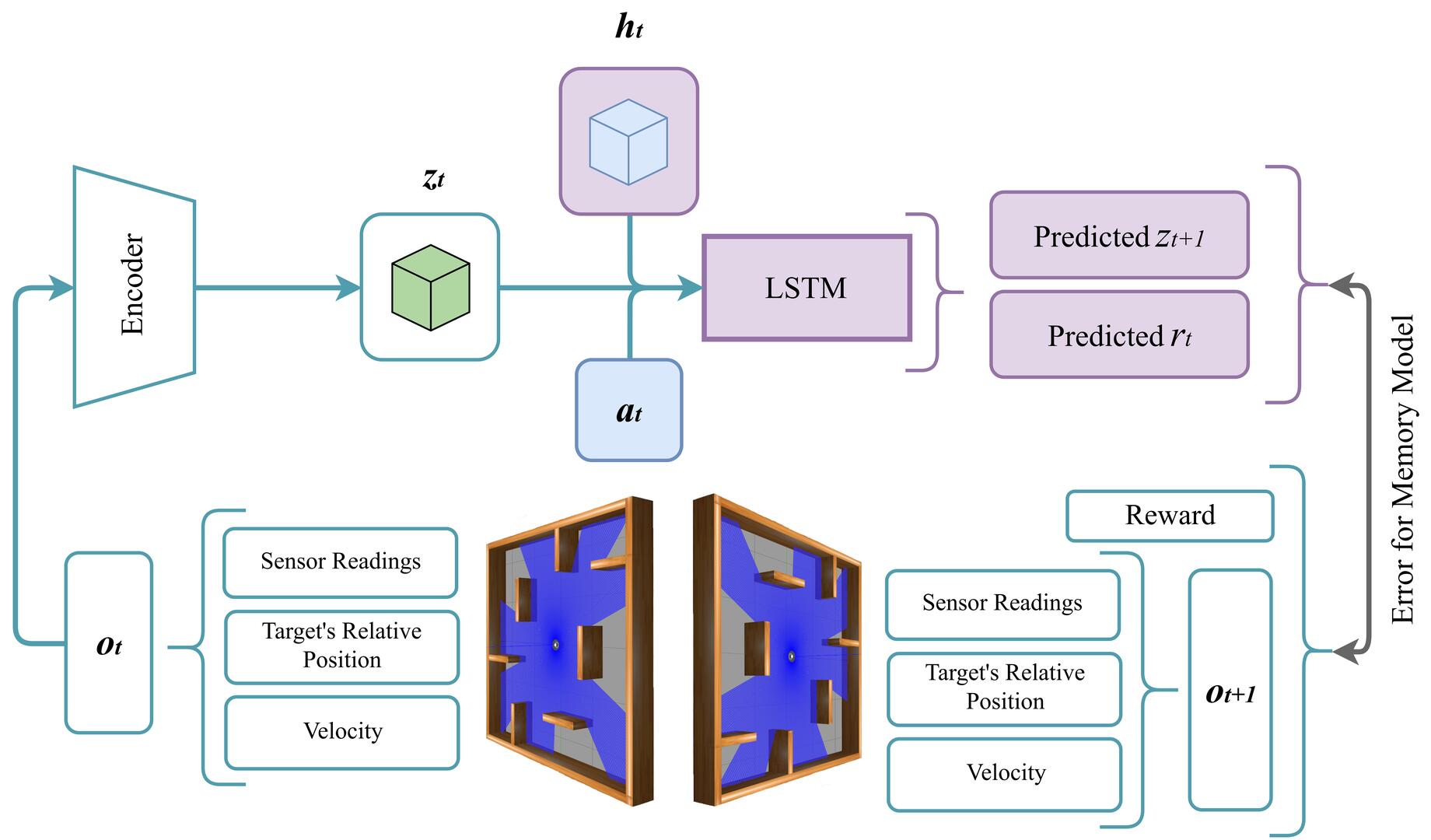}
    \caption{Memory model learning process diagram.}
    \label{fig:mem_model_learning}
\end{figure}

The Memory model employs an LSTM to capture temporal dynamics. It predicts subsequent latent vectors, rewards, and environment flags using the current latent vector \( z_t \), action \( a_t \), and its internal hidden state \( h_t \). The prediction error between the generated predictions and actual outcomes trains this model, illustrated in Fig.~\ref{fig:mem_model_learning}.

\begin{figure}[bp!]
    \centering
    \includegraphics[width=\columnwidth]{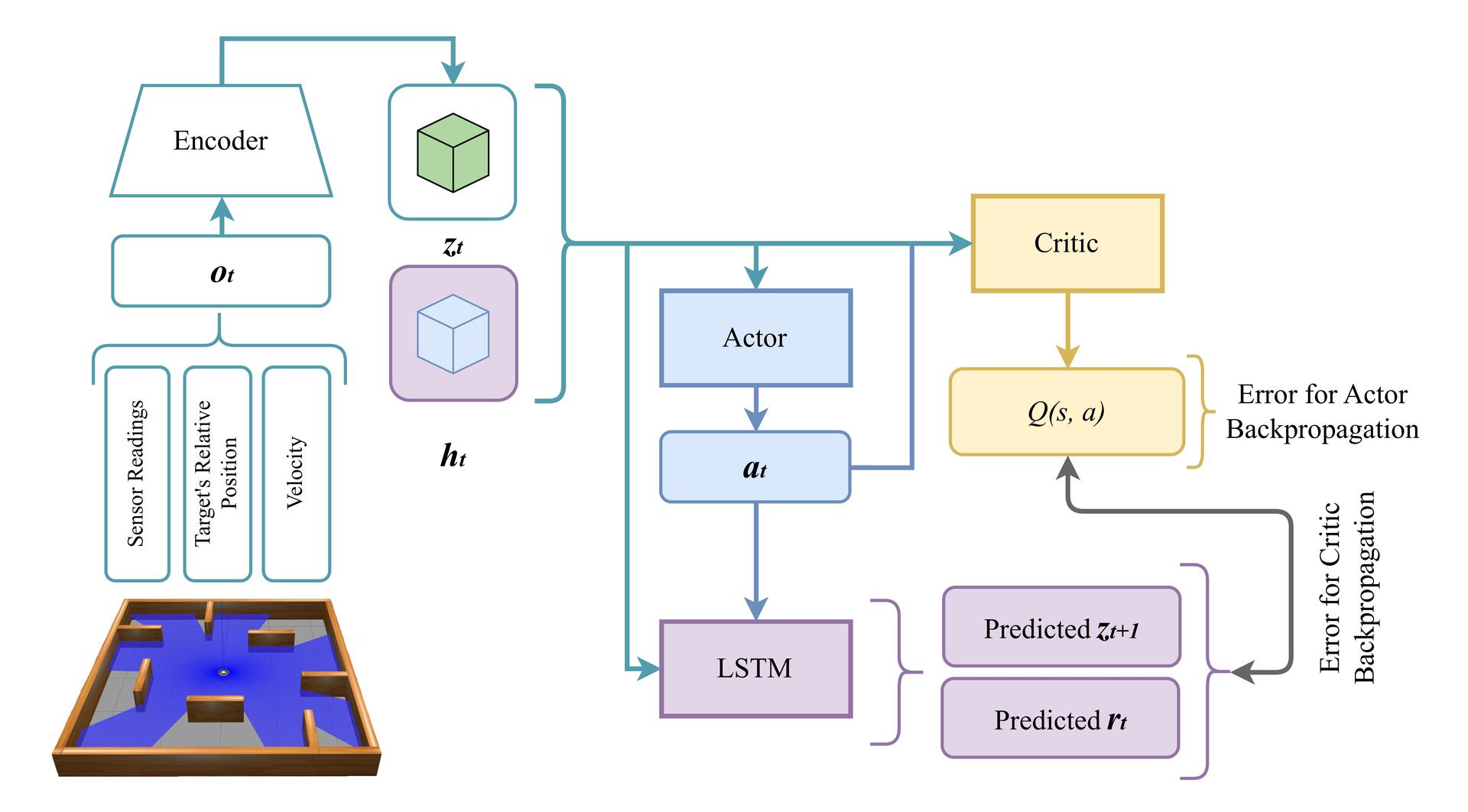}
    \caption{Controller learning process diagram.}
    \label{fig:controller_learning}
\end{figure}

The Actor-Critic network determines actions based on the latent state \( z_t \) and hidden state \( h_t \). The critic evaluates action values \( Q(s, a) \) to guide actor optimization. Targets for critic error calculations utilize future states predicted by the memory model. Training involves simulated rollouts from the vision and memory models, shown in Fig.~\ref{fig:controller_learning}.

\subsection{Hyperparameters}

We implemented SAC, DDPG, and TD3 following their original designs~\cite{haarnoja2018soft,lillicrap2015continuous,fujimoto2018addressing}. Networks consisted of two fully connected layers (400 and 300 units) with ReLU activation. All models were fine-tuned.

The DreamerV3 model~\cite{hafner2023mastering} utilized a slightly reduced architecture compared to its original version to enhance computational efficiency and training stability. Specifically, we adopted an MLP-based encoder-decoder due to the absence of visual input, and reduced network layer sizes by half. The specific hyperparameters for our architecture can be found in the Supplementary Material, since there is an extensive set of hyperparameters, and most of them equal the original implementation~\cite{hafner2023mastering}.

Hyperparameters for each model-free algorithm, summarized in Table~\ref{tab:hyperparams}, were chosen based on standard practices and fine-tuning. Table~\ref{tab:parameters} reports the number of trainable parameters.

\begin{table}[h!]
\small
\centering
\caption{Hyperparameters used for each model-free algorithm.}
\label{tab:hyperparams}
\resizebox{0.95\columnwidth}{!}{%
\begin{tabular}{lccc}
\toprule
\textbf{Hyperparameter} & \textbf{SAC} & \textbf{DDPG} & \textbf{TD3} \\
\midrule
Learning Rate (Actor, Critic) & 0.0003, 0.0003 & 0.0001, 0.001 & 0.001, 0.001 \\
Discount Factor ($\gamma$) & 0.99 & 0.99 & 0.99 \\
Soft Update Coefficient ($\tau$) & 0.001 & 0.001 & 0.005 \\
Replay Buffer Size & 100{,}000 & 100{,}000 & 100{,}000 \\
Batch Size & 128 & 128 & 128 \\
Actor/Critic Layer Sizes & [400, 300] & [400, 300] & [400, 300] \\
Reward Scale & 2 & — & — \\
Noise (TD3 Only) & — & — & 0.1 \\
Update Actor Interval (TD3) & — & — & 2 \\
\bottomrule
\end{tabular}
}
\end{table}

\begin{table}[h!]
\small
\centering
\caption{Total number of trainable parameters given the number of distance readings.}
\label{tab:parameters}
\resizebox{0.92\columnwidth}{!}{%
\begin{tabular}{lrr}
\toprule
\textbf{Algorithm} & \textbf{10 Readings} & \textbf{360 Readings} \\
\midrule
DreamerV3~\cite{hafner2023mastering} & 4{,}163{,}089 & 4{,}521{,}839 \\
SAC~\cite{haarnoja2018soft} & 635{,}508 & 1{,}335{,}508 \\
DDPG~\cite{lillicrap2015continuous} & 514{,}406 & 1{,}074{,}406 \\
TD3~\cite{fujimoto2018addressing} & 763{,}408 & 1{,}603{,}408 \\
\bottomrule
\end{tabular}
}
\end{table}

Given the modifications (ROS2 upgrade and controlled step rates) and the use of full sensor arrays previously untested in the literature, all algorithms were retrained from scratch to ensure fairness and validity in comparative evaluation; no previous results were considered.

%% file: sessions/5_results.tex
\section{Results}\label{results}

In this section, we present and discuss the results obtained from training and evaluating the algorithms across different navigation stages. Each algorithm was trained for 5000 episodes per stage and evaluated over 100 episodes. For the initial four stages, training began from scratch. However, for the more challenging stages 5 and 6, characterized by larger environments and reduced success probability under random policies, we utilized pre-trained models to facilitate learning. Specifically, models were pre-trained on stage 4 before initiating stage 5 training, and subsequently, models trained on stages 4 and 5 were used as a starting point for training stage 6.

We conducted experiments under two distinct sensor input configurations: a reduced set of 10 LIDAR readings, common in prior studies~\cite{de2021soft}, and the complete 360-readings sensor data. The 360-reading configuration, though more computationally demanding, provides a richer spatial context, enhancing decision-making and environmental awareness.

The \textit{success rate}, detailed in Tables~\ref{test_10} and~\ref{test_360}, denotes the proportion of evaluation episodes in which the robot successfully navigated to the target location without collision or exceeding the step limit.

\subsection{Performance with Reduced Sensor Input (10 Readings)}

\begin{figure}[tbp]
    \centering
    \includegraphics[width=\columnwidth]{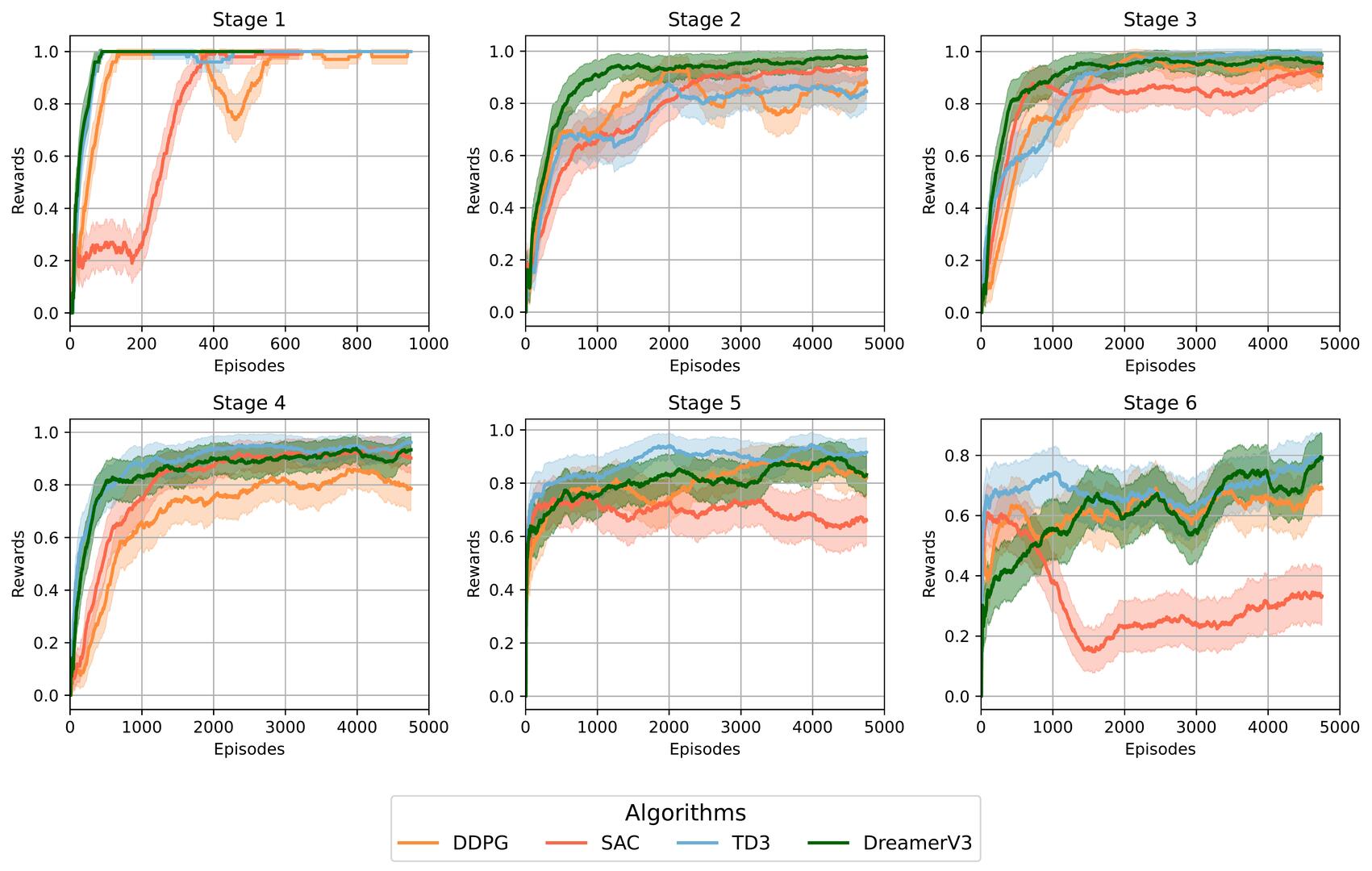}
    \caption{Learning curves for agents trained with 10 sensor readings.}
    \label{learning_curve_10}
\end{figure}

Fig.~\ref{learning_curve_10} illustrates the learning curves for the reduced sensor configuration. All evaluated algorithms achieved successful learning, with DreamerV3 notably displaying a more stable and consistent improvement in rewards throughout training. Testing results summarized in Table~\ref{test_10} demonstrate that DreamerV3 outperformed baseline methods in stages 2, 4, and 6, achieving the highest mean and median success rates across all environments.

\subsection{Performance with Full Sensor Input (360 Readings)}

\begin{figure}[tbp]
    \centering
    \includegraphics[width=\columnwidth]{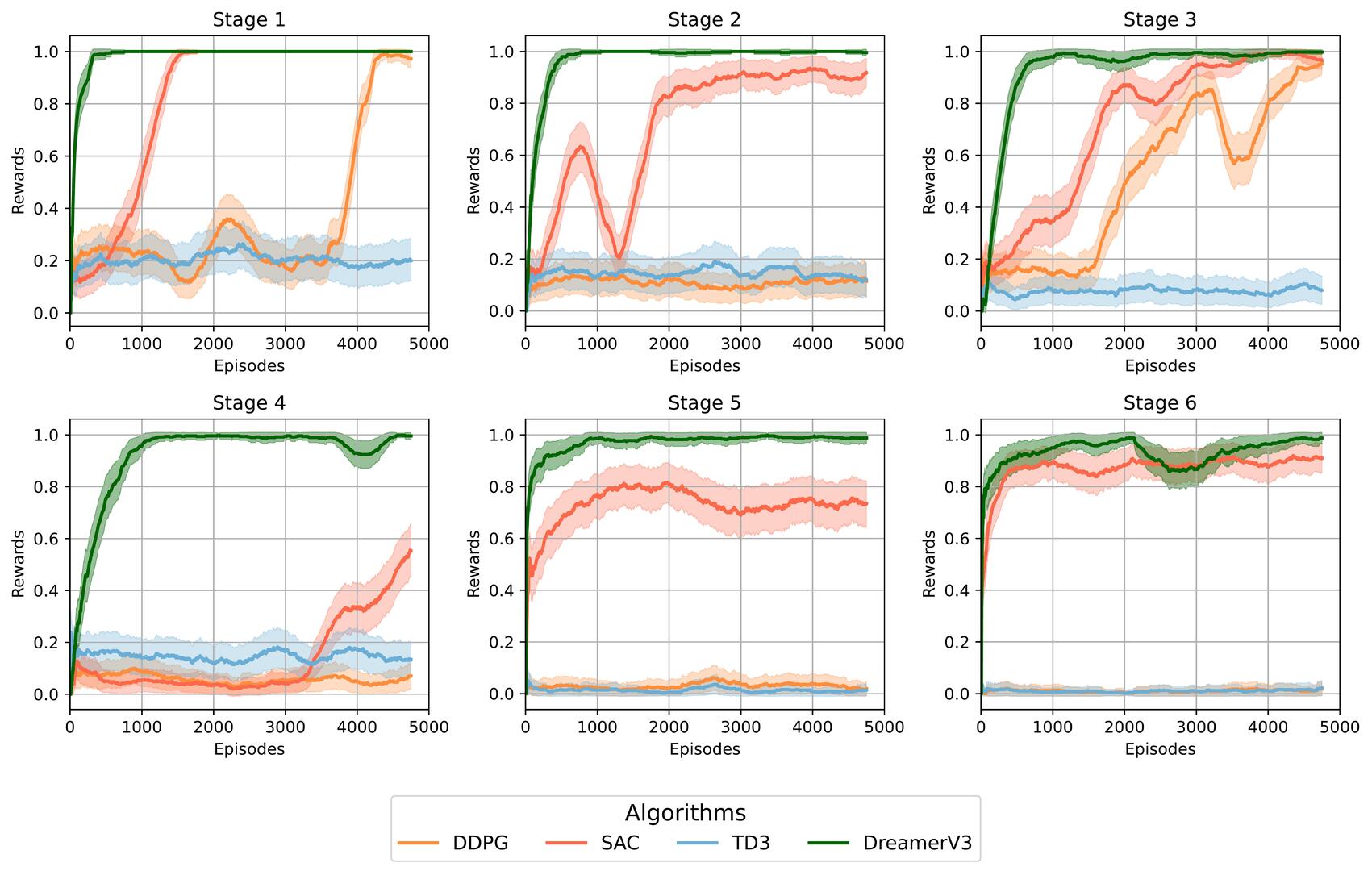}
    \caption{Learning curves for agents trained with 360 sensor readings.}
    \label{learning_curve_360}
\end{figure}

Fig.~\ref{learning_curve_360} shows learning curves for the complete LIDAR data input. DreamerV3 successfully adapted to this increased input dimensionality, demonstrating rapid and stable learning. Conversely, the baseline algorithms encountered significant difficulties. SAC exhibited unstable and suboptimal convergence, DDPG showed limited learning capabilities restricted mostly to simpler environments (stages 1 and 3), and TD3 consistently failed to learn effective policies.

These observations are reinforced by test results reported in Table~\ref{test_360}, where DreamerV3 achieved a perfect 100\% success rate across all stages, markedly outperforming other methods, which struggled significantly under the 360-reading configuration. This clearly highlights DreamerV3’s robustness and scalability, effectively exploiting the rich spatial data provided by complete sensor coverage to enhance obstacle avoidance and route planning.

\subsection{Analysis and Discussion}

DreamerV3 consistently demonstrated superior performance, particularly in challenging scenarios involving long-distance navigation (stage 6) and densely cluttered environments (stage 4). The improved outcomes from using 360-degree sensor data underscore the advantages of richer environmental perception, directly benefiting path planning and obstacle avoidance.

It is important to emphasize that the performance deterioration observed in baseline algorithms (DDPG, SAC, and TD3) with full 360-degree sensor input aligns with findings from previous studies. Such model-free algorithms, initially designed for reduced sensor inputs, lack the scalability to efficiently process and learn from high-dimensional observations. The limitations in their simpler network architectures lead to inefficient feature extraction and unstable learning under increased sensory complexity. These results reinforce the importance of adopting scalable, model-based architectures, such as DreamerV3, which can effectively handle extensive sensor inputs without compromising performance.

\begin{table}[h!]
\small
\centering
\caption{Success rate (\%) across six stages (10 distance sensor readings). Algorithm references: DDPG~\cite{lillicrap2015continuous}, SAC~\cite{haarnoja2018soft}, TD3~\cite{fujimoto2018addressing}, DreamerV3~\cite{hafner2023mastering}.}
\label{test_10}
\resizebox{0.92\columnwidth}{!}{%
\begin{tabular}{lrrrr}
\toprule
\textbf{Stage} & \textbf{DDPG} & \textbf{SAC} & \textbf{TD3} & \textbf{DreamerV3} \\
\midrule
1 & \textbf{100.00} & 99.00 & \textbf{100.00} & \textbf{100.00} \\
2 & 77.00 & 89.00 & 85.00 & \textbf{99.00} \\
3 & 99.00 & 96.00 & \textbf{100.00} & 99.00 \\
4 & 89.00 & 95.00 & 94.00 & \textbf{96.00} \\
5 & 90.00 & 84.00 & \textbf{93.00} & 87.00 \\
6 & 71.00 & 56.00 & 75.00 & \textbf{80.00} \\
\midrule
\textbf{Mean} & 87.67 & 86.50 & 91.17 & \textbf{93.50} \\
\textbf{Median} & 89.50 & 92.00 & 93.50 & \textbf{97.50} \\
\bottomrule
\end{tabular}
}
\end{table}

\begin{table}[h!]
\small
\centering
\caption{Success rate (\%) across six stages (360 distance sensor readings). Algorithm references: DDPG~\cite{lillicrap2015continuous}, SAC~\cite{haarnoja2018soft}, TD3~\cite{fujimoto2018addressing}, DreamerV3~\cite{hafner2023mastering}.}
\label{test_360}
\resizebox{0.92\columnwidth}{!}{%
\begin{tabularx}{\columnwidth}{l *{4}{>{\centering\arraybackslash}X}}
\toprule
\textbf{Stage} & \textbf{DDPG} & \textbf{SAC} & \textbf{TD3} & \textbf{DreamerV3} \\
\midrule
1 & 96.00 & \textbf{100.00} & 18.00 & \textbf{100.00} \\
2 & 12.00 & 80.00 & 17.00 & \textbf{100.00} \\
3 & \textbf{100.00} & \textbf{100.00} & 7.00 & \textbf{100.00} \\
4 & 10.00 & 54.00 & 19.00 & \textbf{100.00} \\
5 & 4.00 & 79.00 & 4.00 & \textbf{100.00} \\
6 & 0.00 & 92.00 & 1.00 & \textbf{100.00} \\
\midrule
\textbf{Mean} & 37.00 & 84.17 & 11.00 & \textbf{100.00} \\
\textbf{Median} & 11.00 & 86.00 & 12.00 & \textbf{100.00} \\
\bottomrule
\end{tabularx}
}
\end{table}

\subsection{Limitations}

While the results presented demonstrate the effectiveness and scalability of the proposed architecture, limitations should be acknowledged.

First, the DreamerV3 architecture, due to its model-based nature and use of larger networks, involves a significantly larger number of hyperparameters and network components compared to conventional model-free algorithms such as DDPG, SAC, or TD3. This increased complexity can lead to higher computational demands during both training and inference. However, it is important to note that modern embedded platforms, such as the NVIDIA Jetson Nano~\cite{nvidia_jetson_nano} and Xavier~\cite{nvidia_jetson_agx_xavier}, are well-suited for running such architectures efficiently in real-time, making practical deployment feasible.

Second, although the proposed method achieves strong performance in simulation across both low and high-dimensional sensor inputs, it has not yet been validated in real-world robotic platforms. This remains an essential step toward verifying robustness under realistic uncertainties, hardware limitations, and sensor noise. We plan to address this in future research by implementing and testing the architecture on a physical TurtleBot3 Burger platform.

Moreover, due to hardware constraints associated with the available computational resources, we were unable to perform multiple runs with different random seeds to provide a statistically robust analysis of the results. Consequently, the reported performance metrics should be interpreted as representative rather than exhaustive indicators of the algorithm’s behavior. We encourage future studies to replicate these experiments under varying initialization seeds to assess variance and further validate the consistency of the observed performance.

Finally, while our current setup includes static obstacle scenarios (as shown in Fig.~\ref{stages}), we have not yet incorporated dynamic or unknown moving obstacles such as other agents or pedestrians. These scenarios are common in real-world environments and present additional challenges for perception and planning. As part of future research, we intend to extend the simulation environments to include such dynamics and evaluate the adaptability of our architecture under these more complex conditions.

%% file: sessions/6_conclusion.tex
\section{Conclusion}\label{conc}

This study introduced a model-based deep reinforcement learning framework for autonomous robot navigation from LIDAR observations. Built upon the DreamerV3 algorithm, the proposed approach integrates a Multi-Layer Perceptron Variational Autoencoder to encode high-dimensional sensor data into compact latent representations, enabling efficient world-model learning and imagination-based policy optimization.

The experimental results demonstrate that our method significantly outperforms model-free baselines from previous works, namely,  SAC~\cite{de2021soft}, DDPG~\cite{grando2022deterministic}, and TD3~\cite{li2022path} across a range of simulated TurtleBot3 navigation environments. In particular, the proposed architecture successfully handled full 360-readings LIDAR inputs, achieving a 100\% success rate in all test scenarios, while model-free methods failed to scale effectively under this conditions. These findings highlight the potential of model-based DRL and unsupervised observation processing in this context.

Future work will focus on extending the proposed framework to more complex and realistic conditions. Specifically, we plan to:
\begin{itemize}
    \item Conduct real-world experiments on physical robotic platforms to validate performance under real-world noise and uncertainty.
    \item Incorporate dynamic and unpredictable obstacles to test adaptability in non-static environments.
    \item Explore multimodal sensory integration, including visual and proprioceptive inputs, to enhance environmental understanding.
    \item Generalize the approach to other robotic domains such as aerial and underwater systems, broadening its applicability to diverse control and perception challenges.
    
\end{itemize}

Overall, this research demonstrates the potential of predictive world models for high-dimensional perception and robust decision-making, advancing the capabilities of autonomous robots operating in complex real-world environments.